\pdfoutput=1
\documentclass[11pt]{article}

\usepackage[final]{acl}

\usepackage{times}
\usepackage{latexsym}
\usepackage{booktabs}
\usepackage{makecell}
\usepackage{xspace}
\usepackage{pifont}
\usepackage{listings}
\usepackage{subfigure}

\usepackage[T1]{fontenc}
\usepackage[utf8]{inputenc}

\usepackage{microtype}

\usepackage{inconsolata}

\usepackage{graphicx}

\definecolor{lightergray}{RGB}{230,230,230}
\definecolor{DarkGreen}{RGB}{30,130,30}
\newcommand{\cmark}{\textcolor{DarkGreen}{\ding{51}}}
\newcommand{\xmark}{\textcolor{red}{\ding{55}}}
\newcommand\ourmethod{UAR\xspace}

\title{Unified Active Retrieval for Retrieval Augmented Generation}

\author{
Qinyuan Cheng$^{1,2}$\thanks{Equal contribution} \hspace{.3em}
Xiaonan Li$^{1}$\footnotemark[1] \hspace{.1em}
Shimin Li$^{1}$ \hspace{.1em}
Qin Zhu$^{1}$ \hspace{.1em}
Zhangyue Yin$^{1}$\\
\textbf{
Yunfan Shao$^{1,2}$ \hspace{.1em}
Linyang Li$^{2}$ \hspace{.1em}
Tianxiang Sun$^{1}$ \hspace{.1em}
Hang Yan$^{2}$ \hspace{.1em}
Xipeng Qiu$^{1,3,}$\thanks{Corresponding author.}}
\\
[1ex]
$^{1}$Fudan University \\
$^{2}$Shanghai AI Laboratory \\
$^{3}$Shanghai Collaborative Innovation Center of Intelligent Visual Computing \\
\tt{chengqy21@m.fudan.edu.cn}
\tt{\{lixn20, xpqiu\}@fudan.edu.cn}
}

\begin{document}
\maketitle
\begin{abstract}
In Retrieval-Augmented Generation (RAG),
retrieval is not always helpful and applying it to every instruction is sub-optimal.
Therefore, determining whether to retrieve is crucial for RAG, which is usually referred to as Active Retrieval. However, existing active retrieval methods face two challenges: 1. They usually rely on a single criterion, which struggles with handling various types of instructions. 2. They depend on specialized and highly differentiated procedures, and thus combining them makes the RAG system more complicated and leads to higher response latency. To address these challenges, we propose \textbf{U}nified \textbf{A}ctive \textbf{R}etrieval (\textbf{UAR}). UAR contains four orthogonal criteria and casts them into plug-and-play classification tasks, which achieves multifaceted retrieval timing judgements with negligible extra inference cost. 
We further introduce the Unified Active Retrieval Criteria (UAR-Criteria), designed to process diverse active retrieval scenarios through a standardized procedure.
Experiments on four representative types of user instructions show that UAR significantly outperforms existing work on the retrieval timing judgement and the performance of downstream tasks, which shows the effectiveness of UAR and its helpfulness to downstream tasks.
\end{abstract}

\section{Introduction}
With the rapid development of large language models (LLMs)~\cite{gpt3, llama2, GLM130B, baichuan, InternLM, Qwen}, AI assistants based on LLMs become unbiquitous and show remarkable abilities on various types of instructions, e.g., coding, writing and reasoning~\cite{ChatGPT, alpaca, vicuna2023, MOSS, gpt4, Claude, Gemini}.
However, LLMs often generate fabricated and non-factual information~\cite{TruthfulQA, HalluQA,Factual_survey}, which is called ``hallucination'' and makes LLMs' responses not trustworthy in real-world scenarios.

Retrieval-Augmented Generation (RAG) is a prevailing approach to address LLM's hallucination~\citep{realm_rag, rag_survey}. Given a user query, it usually first retrieves relevant documents and then uses them to augment the LLM's factual correctness.
However, retrieval is not always helpful and applying it to every instruction is sub-optimal.
When faced with instructions that do not require external knowledge, RAG can impair the creativity and versatility of LLMs~\cite{Self-RAG}.
\begin{figure}[t]
\centering
  \includegraphics[width=0.9\columnwidth]{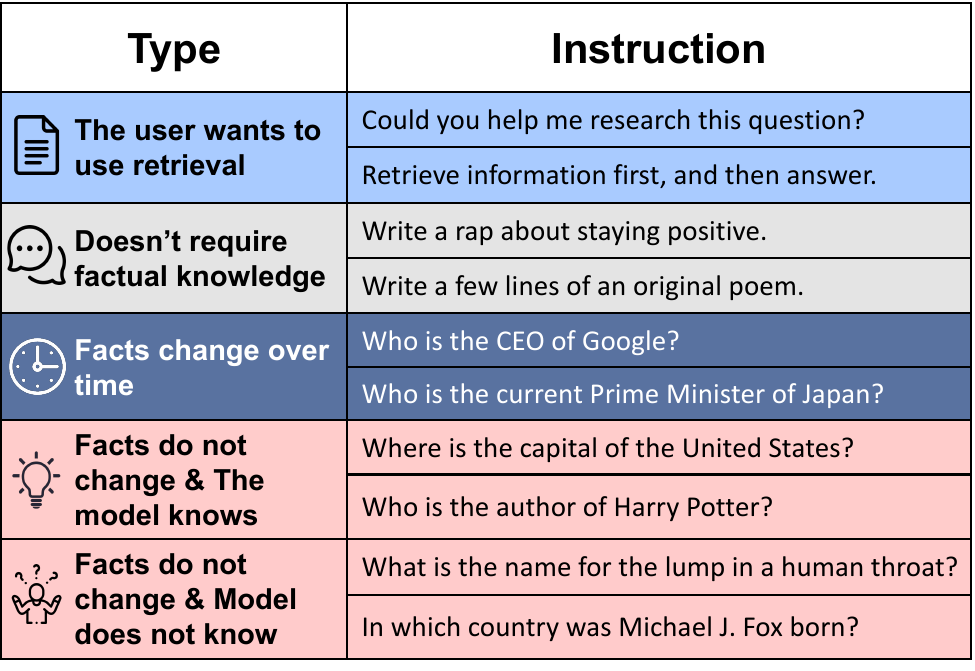}
  \caption{Different types of user instructions, which can not be handled by single active retrieval criteria.}
  \label{fig:question_examples}
  \vskip -0.2in
\end{figure}

If irrelevant knowledge is retrieved, it will hinder the LLM from utilizing its internal knowledge effectively and make it produce low-quality responses~\cite{Irrelevant_context, robust_to_irrelevant_document}.
Meanwhile, compared with only LLM, RAG involves an additional retrieval process and the longer LLM input,
resulting in significantly longer response latency.
Therefore, applying RAG for all instructions is sub-optimal and unnecessary, and determining the correct timing for retrieval is crucial for LLMs' real-world application, which is often reftered to as \textit{Active Retrieval}~\citep{Flare,Self-RAG}.

\begin{table*}[!t]
    \centering
    \small
    % \resizebox{0.95\linewidth}{!}{
    \begin{tabular}{lcccc}
        \toprule
         & \makecell{\textbf{\ourmethod} \\ (our work)} &
         \makecell{ \textbf{FLARE} \\ \citep{Flare} } & 
        \makecell{ \textbf{Self-RAG} \\ \citep{Self-RAG} } &  
        \makecell{ \textbf{SKR} \\ \citep{SKR} }
        \\
         \cmidrule(lr){1-1}  \cmidrule(lr){2-2}  \cmidrule(lr){3-3}  \cmidrule(lr){4-4}  \cmidrule(lr){5-5} 
          Intent Awareness?   & \cmark & \xmark & \xmark & \xmark  \\ 
          Knowledge Awareness? & \cmark & \xmark & \cmark & \xmark   \\ 
          Time Awareness?   & \cmark & \xmark & \xmark & \xmark   \\ 
          Self Awareness?   & \cmark & \cmark & \xmark & \cmark \\ 
         \bottomrule
    \end{tabular}
    % }
    \caption{
    Comparison of UAR to other active retrieval methods. Exciting methods only consider a single active retrieval criterion, while UAR unifies four orthogonal criteria and can handle various types of user instructions.
    }
    \label{tab:method_comparison}
    \vskip -0.1in
\end{table*}

In general, there are two lines of active retrieval methods. One is the ``knowledge-aware'' method, based on the instruction's factual relevance, e.g., Self-RAG~\citep{Self-RAG}. If the instruction requires factual information, the retrieval will be triggered. Another line of work is the ``self-aware'' method,
based on the LLM's self awareness~\citep{SKR}. The retrieval is only triggered when the LLM thinks that it does not know the answer, i.e., when it is uncertain. In this way, the retrieval can supplement knowledge for the LLM when necessary and avoid unnecessary retrieval cost.
Although these methods can determine retrieval timing for specialized scenarios, they still face two challenges: 
1. Previous work usually relies on a single criterion, which struggles with diverse scenarios.
For instance, the self-aware method~\citep{SKR, RA-ISF, rowen} struggles with various instructions such as time-sensitive queries or those with user's explicit retrieval intent.
For time-sensitive questions, it is challenging for a static LLM to judge whether it possesses the correct knowledge for a rapidly changing answer.
Additionally, these methods often overlook user's intent in real-world scenarios, such as when a user seeks a verifiable answer that requires external information sources, necessitating retrieval.
Therefore, correctly determining whether to retrieve requires multifaceted decision-making. 
2. Existing methods rely on specialized procedures, complicating the integration within the RAG system and increasing computational load.
For example, FLARE~\citep{Flare} uses the confidence of generation and Rowen~\citep{rowen} relies on response divergence for the same question. 
These highly differentiated approaches are difficult to unify, making it very difficult to extend them to new scenarios.

To address these challenges, we propose \textbf{U}nified \textbf{A}ctive \textbf{R}etrieval (\textbf{UAR}), a unified and comprehensive framework for judging whether to retrieve for various types of user instructions. 
UAR consists of various orthogonal criteria of retrieval timing and casts them into unified classification tasks, and thus can judge the LLM's retrieval timing both comprehensively and efficiently.
Specifically, UAR consists of four orthogonal criteria for determining the retrieval timing: 
1) \textbf{Intent-aware}: whether the user desires retrieval / external information;
2) \textbf{Knowledge-aware}: whether the question requires fact knowledge;
3) \textbf{Time-Sensitive-aware}: whether the question is time-sensitive;
4) \textbf{Self-aware}: whether the LLM has the internal knowledge.
As shown in Table~\ref{tab:method_comparison}, compared with previous methods of single criterion~\citep{Flare,SKR,Self-RAG},
UAR can comprehensively handle various types of user instructions and call retrieval accurately considering multiple active retrieval criteria.
To efficiently achieve judgements of multiple criteria, UAR unifies each criterion's judgement into binary classification tasks using lightweight classifiers.
For each criterion $c_i$, we train a plug-and-play binary classifier on the last layer's hidden states of a fixed LLM, to judge whether the input requires retrieval according to $c_i$.
In this way, UAR does not change LLMs' parameters, avoiding the costly LLM fine-tuning and performance degradation~\citep{sft_degradation}.
Meanwhile, the classifiers and LLM generation share the same input encoding, which makes UAR only need to encode the input once and thus achieves multifaceted retrieval timing judgements with negligible extra inference cost.

To handle various instructions in an unified procedure, we further propose Unified Active Retrieval Criteria (UAR-Criteria), which specifies priorities for multiple retrieval criteria and unifies them into a single multifaceted decision tree. 
As shown in Figure~\ref{fig:UAR_framework}, UAR-Criteria can trigger retrieval for time-sensitive or LLM-unknown instructions, which facilitates necessary external information supplement. Meanwhile, UAR-Criteria cancels retrieval for those non-knowledge-intensive or LLM-known instructions, which avoids the negative effect of unnecessary retrieval.
In this way, UAR-Criteria unifies the process to comprehensively decide whether to retrieval for various types of user instructions, which facilitates more effective RAG.

Experiments on four representative types of user instructions show that UAR significantly outperforms existing work on the retrieval timing judgement accuracy and the performance of downstream tasks, which verifies the effectiveness of UAR and its helpfulness to downstream tasks.
We summarize our contributions as follows: 
\begin{itemize}
    \item We propose an active retrieval framework named Unified Active Retrieval (UAR) for Retrieval-Augmented Generation (RAG). To the best of our knowledge, UAR is the first work to propose multifaceted criteria for active retrieval and demonstrate its necessity. 
    \item We curate the Active Retrieval benchmark (AR-Bench) for evaluating the accuracy of retrieval timing and conduct comprehensive experiments on AR-Bench and downstream tasks. The results show that UAR significantly outperforms existing work and achieves more efficient RAG.
    \item We release the code, data, models and relevant resources to facilitate future research\footnote{\url{https://github.com/xiami2019/UAR}}.
\end{itemize}

\section{Related Work}

\subsection{Active Retrieval}
Compared to applying retrieval for every instruction (passive retrieval), active retrieval has advantages such as not hurting the versatility of the model, reducing the number of retrievals, and preventing interference from low-quality retrieval results.
Self-RAG \citep{Self-RAG} construct active retrieval data using GPT-4 and teach the model to not retrieve when encounter non-knowledge-intensive instructions.
FLARE \citep{Flare} proposes forward-looking active retrieval augmented generation based on model's confidence, only retrieving information when the model's uncertainty for the prediction is high.
SKR \citep{SKR}, RA-ISF \citep{RA-ISF} and Self-DC \citep{self-dc} first determines whether the model knows the questions and then retrieves only when the model does not know.
However, current active retrieval methods mostly consider only a single scenario and are unable to adapt to complex situations in real-world applications.

\subsection{Time-awareness of LLMs}
There are some papers focus on the time awareness of large language models.
\citet{Time_sensitive_questions} construct a time-sensitive QA dataset called TimeQA to evaluate the model's ability to handle temporal questions.
\citet{Mulan} create a benchmark named MULAN for evaluating the ability of language models to predict mutable facts.
They find representations classification can distinct immutable and mutable facts, which means language models have a certain degree of temporal awareness.
\citet{TAQA} investigate whether language models can align their internal knowledge to a target year.
They construct a dataset which contains time-sensitive questions.

\subsection{Self-awareness of LLMs}
Self-awareness means that large language model can be aware of what they know and what they don't know.
\citet{know_what_they_know} find that language models can be well-calibrated when using a multiple-choice template.
And they also finetune a value head to predict whether language models know the answer to the given question.
\citet{express_uncertainty} finetune GPT-3 to express uncertainty in words on math questions.
\citet{self_aware} collect some unanswerable questions to evaluate whether language models can express uncertainty to these unanswerable questions.
\citet{R-Tuning} utilize supervised fine-tune to teach large language models to refuse questions which beyond their knowledge scope.
\citet{Say_IDK} explore more alignment methods beyond supervised fine-tuning to teach language models know and express what they don't know, like preference optimization.
Results of previous work show that we can enhance language models' self-awareness with corresponding dataset.

\begin{figure*}[t]
\centering
\includegraphics[width=0.9\linewidth]{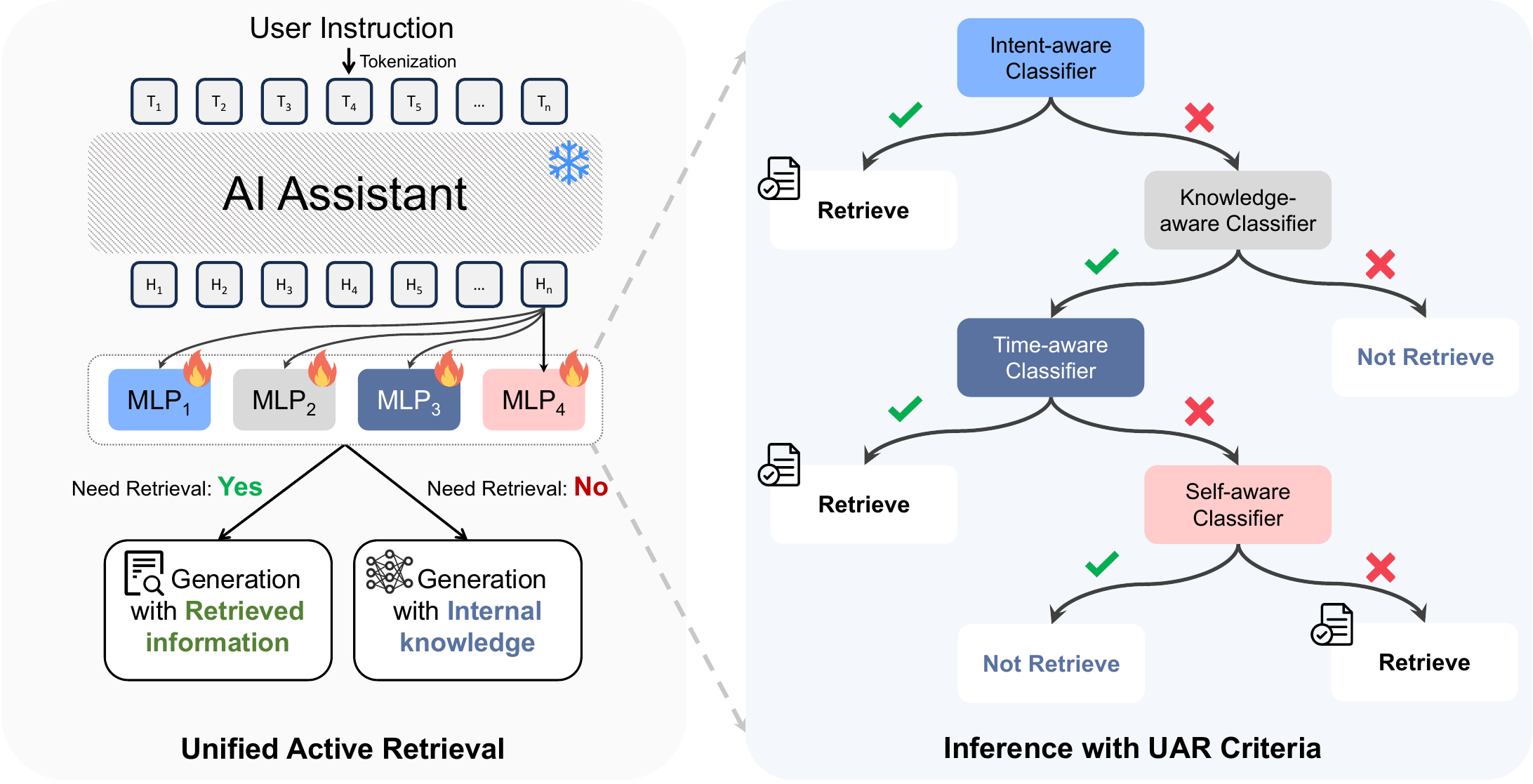}
\caption{Overview of the UAR framework. \includegraphics[height=1em]{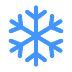} indicates that we freeze these parameters. \includegraphics[height=1em]{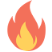} indicates that we update these parameters. Each MLP is a fully connected layer, with an input dimension equal to the model's hidden state dimension and an output dimension of 2.}
\label{fig:UAR_framework}
\vskip -0.1in
\end{figure*}

\section{Methodology}
UAR is a plug-and-play active retrieval framework.
As shown in Figure \ref{fig:UAR_framework}, we fix the parameters of the LLM and train a lightweight classifier for each active retrieval criteria using the model's hidden states, which is far more efficient than fine-tuning the entire model.
Besides, UAR determines the need for active retrieval following the UAR-Criteria shown on the right side of Figure \ref{fig:UAR_framework}, invoking retrieval when necessary and avoiding unnecessary across various scenarios, making RAG more effective and efficient.
For instructions requiring retrieval, we append the retrieved documents to the original instruction, 
which means that UAR does not introduce extra LLM inference cost.
We introduce the details of our UAR framework in the following sections.

\subsection{UAR Classifiers Training}
\label{sec:cls_training}
We construct distinct training data tailored to each scenario.

\paragraph{Self-aware} In the self-aware scenario, the model must determine if it knows the answer to a given question.
Following the methodology in \citet{Say_IDK}, we create model-specific IDK (I don't know) datasets.
For example, with the Llama2-7B-chat model, we use the TriviaQA \citep{TriviaQA} dataset, sampling ten responses for each question.
If all responses are correct, the question is marked as known; otherwise, it is unknown.
10\% of the TriviaQA training set is used for validation, with the rest designated as the training set.

\paragraph{Time-aware} In the time-aware scenario, it is critical to determine if a user's question is time-sensitive, meaning the answer changes over time.
We utilize questions from TAQA's \citep{TAQA} training and validation sets as time-sensitive questions.
In contrast, we sample an equivalent number of questions from the TriviaQA training set to represent non-time-sensitive questions, which typically have static answers.

\paragraph{Knowledge-aware} In the knowledge-aware scenario, identifying whether a user's instruction requires factual knowledge is essential. 
We use non-retrieval instruct-following data from the Self-RAG \citep{Self-RAG} training set, which GPT-4 classifies as non-knowledge-intensive.
We select 2,000 entries for our validation set and 22,801 for training.
Additionally, we incorporate all entries from our time-aware data's training and validation sets as knowledge-intensive instructions to complete the final knowledge-aware training and validation sets.

\paragraph{Intent-aware} In the intent-aware scenario, it's crucial to identify users' intentions to use retrieval-augmented generation.
Due to a lack of data with explicit retrieval intentions, we use Self-Instruct \citep{self_instruct} to generate 3,000 user intents from ten handwritten intents.
We allocate 2,000 for training, 500 for validation, and 500 for testing.
We assemble user queries by sampling 52,949 entries from Self-RAG's non-retrieval-required data, and factual knowledge questions from TAQA and TriviaQA for the training set, with an additional 5,000 for validation.
We integrate half of these data with user retrieval intents, alternating the position of intents before and after user inputs, to create inputs with retrieval intents. The remaining data are used as inputs without retrieval intents.

For each scenario, we train a single-layer MLP as the classifier, using the hidden states from the last token in the input as the input to the classification head. In this way, UAR can achieve various criteria's judgements with negligible extra computational cost.
We include details of classifiers' training in Appendix \ref{apx:uar_training}.

\subsection{UAR Criteria}
We further propose UAR-Criteria to unify the judgements of different types of user instructions in to one unified procedure.
During the inference stage, UAR sequentially utilizes four classifiers according to different priorities to determine the correct timing for retrieval calls, and we introduce its details as follows.

Initially, UAR checks whether the user intends to use retrieval augmentation.
If so, retrieval is triggered.
If not, UAR evaluates whether the input is knowledge-intensive.
For non-knowledge-intensive tasks, retrieval is not used.
For knowledge-intensive tasks, UAR further assesses whether the knowledge is time-sensitive.
Retrieval is necessary for time-sensitive questions.
For non-time-sensitive, knowledge-intensive tasks, UAR checks whether the model already has the relevant knowledge, activating retrieval only for unfamiliar questions.
In this way, UAR can handle various types of instructions. 
Specifically, UAR-Criteria activates retrieval for instructions that are time-sensitive, unknown to the model, and have explicit retrieval intent, which facilitates necessary external information supplement.
Meanwhile, UAR-Criteria cancels retrieval for those non-knowledge-intensive or LLM-known instructions, which avoids the negative effect of unnecessary retrieval. 
Meanwhile, since UAR achieves the judgement of multifaceted criteria by linear classifiers, the introduced extra computational cost is negligible.

\subsection{Generation with Relevant Information}
For instructions requiring retrieval augmentation, we append the retrieved external information with a RAG template to the original user input. Since most of the prevailing LLMs are based on the decoder-only architecture~\citep{gpt3}, UAR can avoid the need to recompute the original instruction.
The retriever might fetch information irrelevant to the question, our prompt instructs the model to utilize only the information relevant to the question.
This approach helps prevent irrelevant information from misleading the model.
An example of our RAG prompt is as follows:
{
\begin{lstlisting} [frame=none]
{question}
Here are some additional reference passages:
{reference passages}
You can refer to the content of relevant reference passages to answer the questions.
Now give me the answer.
\end{lstlisting}
}
For instructions that do not require retrieval, we allow the model to generate outputs in its original format.

\section{Experiments}

\subsection{Benchmarking Retrieval Timing}
We curate an Active Retrieval Benchmark (AR-Bench) to evaluate the accuracy of various active retrieval methods in determining the timing of retrieval.
The AR-Bench includes four sub-tasks: intent-aware, knowledge-aware, time-aware and self-aware, covering all the active retrieval scenarios mentioned in this paper.
Each sub-task is a binary classification task comprising 8,000 samples, with a 1:1 ratio of positive to negative examples, and these samples do not overlap with the training data of UAR.
These four sub-tasks separately evaluate one single active retrieval criterion and we control variables to ensure that each task's retrieval decision solely depends on one single criterion.
We introduce details of AR-Bench construction in Appendix \ref{apx:ar_bench_details}.

\begin{table*}[t]
\centering
\small
\setlength\tabcolsep{7pt}
\begin{tabular}{@{}lccccc@{}}
\toprule
\textbf{Scenario} & 
\textbf{Intent-aware} & 
\textbf{Knowledge-aware} &
\textbf{Time-aware} &
\textbf{Self-aware} &
\textbf{Overall} \\
\midrule
\midrule
\multicolumn{6}{c}{\textit{7B Models}} \\
\midrule
FLARE & 61.95 & 56.76 & 53.69 & 53.59 & 56.50 \\
Self-RAG\(^\dagger\) & 64.26 & 72.82 & 47.45 & 55.95 & 60.12  \\
SKR  & 58.73 & 42.94 & 76.61 & 70.28 & 62.14 \\
\midrule
UAR  & \textbf{91.88} & \textbf{90.38} & \textbf{86.69} & \textbf{72.32} & \textbf{85.32} \\
\midrule
\midrule
\multicolumn{6}{c}{\textit{13B Models}} \\
\midrule
FLARE  & 65.49 & 53.54 & 55.20 & 54.61 & 57.21 \\
Self-RAG\(^\dagger\) & 67.80 & 64.85 & 54.44 & 52.49 & 59.89  \\
SKR  & 59.00 & 43.18 & 79.91 & 68.70 & 62.70 \\
\midrule
UAR  & \textbf{92.49} & \textbf{91.04} & \textbf{87.94} & \textbf{73.84} & \textbf{86.33} \\
\bottomrule
\end{tabular}
\caption{Comparisons of active retrieval accuracy on our active retrieval benchmark (AR-Bench). \(\dagger\): Self-RAG is fine-tuned from Llama2-base models. Other methods are based on Llama2-chat models.} 
\label{tab:active_retrieve_acc}
\vskip -0.2in
\end{table*}

\subsection{Downstream Tasks}
We select six datasets to test UAR's performance in real downstream tasks and its adaptability to different active retrieval scenarios.
Since the intent-aware judgement focuses on satisfying users' retrieval intent, which is not reflected on the objective downstream performance, the selected datasets cover the remaining three scenarios: knowledge-aware, time-aware, and self-aware.
For knowledge-aware scenario, we use DROP \citep{DROP} and \citep{GSM8K}.
For time-aware scenario, we use TAQA \citep{TAQA} and FreshQA \citep{FreshQA}.
For self-aware scenario, we use TriviaQA \citep{TriviaQA} and WebQuestions (WQ) \citep{WQ}.
We provide a detailed introduction to these datasets in Appendix \ref{apx:downstream_dataset}.
In these six datasets, we only use the training sets of TriviaQA anf TAQA for UAR's training, and thus the remaining evaluation dataset can reflect the UAR's out-of-distribution (OOD) performance, which can further verify the effectiveness of UAR in complicated real-world scenarios.

\subsection{Baselines}
We choose three active retrieval methods as our baseline methods: FLARE \citep{Flare}, Self-RAG \citep{Self-RAG}, and SKR \citep{SKR}, covering two main active retrieval criteria.
FLARE determines whether external retrieval is needed by assessing the model's uncertainty about the generated responses.
SKR first collects model's self-knowledge (knowns and unknowns) data, then trains a BERT-based \citep{BERT} classifier to determine whether the model knows a certain question.
For questions the model does not know, retrieval augmentation is used.
Self-RAG gathers a large amount of knowledge-intensive and instruction-following data (no fact knowledge required), then trains the pre-trained model to only use retrieval augmentation for knowledge-intensive tasks.
For downstream tasks, we also include generation with never-retrieval and always-retrieval as baseline methods.
The original SKR and FLARE are not based on Llama2, so we re-implement these methods on the Llama2 model.
The details of our re-implementation are provided in Appendix \ref{apx:baseline_implementation}.

\subsection{Retrievers}
For time-sensitive datasets TAQA and FreshQA, we follow the settings in FreshQA \citet{FreshQA} and use Google Search.
For other datasets, following the settings in Self-RAG, we use off-the-shelf Contriever-MS MARCO \citep{Contriever} and retrieve up to ten documents for each input.
During generation, we use the top five retrieved documents.
For other datasets, following the settings in Self-RAG, we adopt off-the-shelf Contriever-MS MARCO~\citep{Contriever} and use the top-5 documents.

\begin{table*}[t]
\small
\centering
\setlength\tabcolsep{4pt}
% \small
\begin{tabular}{@{}lccccccc@{}}
\toprule
\textbf{Dataset} & \textbf{Drop} & \textbf{GSM8K} & \textbf{TriviaQA} & \textbf{WQ} & \textbf{TAQA} & \textbf{FreshQA} & 
\textbf{Overall} \\
\midrule
\midrule
\multicolumn{8}{c}{\textit{7B Models}} \\
\midrule
\midrule
Never-Ret & 57.67$_{(0\%)}$ & 26.91$_{(0\%)}$ & 62.15$_{(0\%)}$ & 59.79$_{(0\%)}$ & 16.43$_{(0\%)}$ & 35.64$_{(0\%)}$ &  43.10 \\
Always-Ret & 49.57$_{(100\%)}$ & 23.65$_{(100\%)}$ & 68.73$_{(100\%)}$ & 53.99$_{(100\%)}$ & 34.49$_{(100\%)}$ & 65.35$_{(100\%)}$ & 49.23  \\
\midrule
\multicolumn{8}{c}{\textit{Active Retrieval}} \\
\midrule
Self-RAG\(^\dagger\) & 39.17$_{(5.7\%)}$ & 16.07$_{(4.9\%)}$ & 61.68$_{(53.5\%)}$ & 43.01$_{(61.9\%)}$ & 11.09$_{(42.1\%)}$ & 44.88$_{(51.2\%)}$ & 35.98 \\
SKR & 53.00$_{(61.4\%)}$ & 26.38$_{(35.3\%)}$ & 65.39$_{(48.9\%)}$ & 58.96$_{(26.8\%)}$ & 30.63$_{(79.9\%)}$ & 48.84$_{(39.3\%)}$ & 47.17 \\
FLARE & \textbf{56.98}$_{(9.6\%)}$ & 26.76$_{(45.8\%)}$ & 65.98$_{(58.8\%)}$ & 55.46$_{(67.9\%)}$ & 28.08$_{(63.5\%)}$ & 57.76$_{(57.4\%)}$ & 48.50 \\
UAR & 52.55$_{(49.7\%)}$ & \textbf{26.91}$_{(0.1\%)}$ & \textbf{69.02}$_{(50.1\%)}$ & \textbf{60.53}$_{(25.0\%)}$ & \textbf{34.46}$_{(99.7\%)}$ & \textbf{59.74}$_{(78.5\%)}$ & \textbf{50.49} \\
\midrule
\midrule
\multicolumn{8}{c}{\textit{13B Models}} \\
\midrule
\midrule
Never-Ret & 58.76$_{(0\%)}$ & 40.64$_{(0\%)}$ & 63.18$_{(0\%)}$ & 57.63$_{(0\%)}$ & 11.14$_{(0\%)}$ & 34.98 $_{(0\%)}$ &  44.39 \\
Always-Ret & 54.16$_{(100\%)}$  & 37.68$_{(100\%)}$  & 71.02$_{(100\%)}$ & 54.08$_{(100\%)}$  & 34.20$_{(100\%)}$  & 62.05$_{(100\%)}$ & 52.09 \\
\midrule
\multicolumn{8}{c}{\textit{Active Retrieval}} \\
\midrule
Self-RAG\(^\dagger\) & 44.68$_{(0.1\%)}$ & 21.00$_{(0.0\%)}$ & 62.53$_{(30.0\%)}$ & 42.37$_{(51.9\%)}$ & 15.42$_{(37.0\%)}$ & 39.60$_{(39.3\%)}$ & 37.60 \\
SKR & 56.58$_{(50.9\%)}$ & 39.35$_{(27.6\%)}$ & 67.21$_{(49.2\%)}$ & 56.20$_{(31.5\%)}$ & 31.66$_{(89.2\%)}$ & 50.17$_{(45.9\%)}$ & 50.16 \\
FLARE & 58.12$_{(17.5\%)}$ & 38.05$_{(61.2\%)}$ & 68.00$_{(54.9\%)}$ & 53.64$_{(69.6\%)}$ & 25.40$_{(60.9\%)}$ & 50.17$_{(55.8\%)}$ & 48.90 \\
UAR & \textbf{58.55}$_{(3.7\%)}$ & \textbf{40.64}$_{(0.0\%)}$ & \textbf{71.71}$_{(48.5\%)}$ & \textbf{59.20}$_{(31.2\%)}$ & \textbf{34.14}$_{(99.6\%)}$ & \textbf{55.45} $_{(73.3\%)}$ & \textbf{53.26} \\
\bottomrule
\end{tabular}
\caption{Comparisons of downstream tasks performance. Never-Ret means that retrieval augmentation is never used during generation, while Always-Ret means that retrieval augmentation is used in every generation.\(\dagger\): Self-RAG is fine-tuned from Llama2-base models. Other methods are based on Llama2-chat models.}
\label{tab:main_result}
\vskip -0.2in
\end{table*}

\subsection{Evaluation Metrics}
Following previous work \citep{Self-RAG, mallen, Toolformer}, we check whether gold answers are included in model's generations to evaluate performance on the DROP, TriviaQA, and WQ datasets, instead of strictly requiring exact matching.
For GSM8K, we use the prompts for answer extraction in \citet{Zero-shot-COT} to extract model's answers and then use exact matching to calculate the accuracy.
For TAQA and FreshQA, since the golden answers are too long to conduct lexical matching, we use ChatGPT to evaluate whether the model's answers are correct.
Details of ChatGPT evaluation are included in Appendix \ref{apx:chatgpt_eval}.
Furthermore, for downstream tasks, we also report the percentage of samples that invoke retrieval.
For AR-Bench, we use accuracy as the metric.
Since AR-Bench is a binary classification task with an equal number of positive and negative samples, accuracy and micro F1 score are equivalent.

\subsection{Comparisons on AR-Bench}
We show the results in Table~\ref{tab:active_retrieve_acc}.
We observe that UAR outperforms existing active retrieval methods across all AR-Bench scenarios, demonstrating its versatility and effectiveness.
Since baseline methods depend on a single criterion, they struggle with various active retrieval scenarios, which demonstrates the limitation of single criterion and the necessity of multifaceted decision for active retrieval.
Additionally, we find FLARE struggle with self-aware scenario, which it is targeted at.
We think it is because its uncertainty estimation heavily depends on model calibration and this leads to its poor performance on less calibrated models like chat models~\citep{aligned_model_calibration} or those with fewer parameters.
Self-RAG uses the knowledge-intensive nature of tasks as the retrieval criterion, performing well in knowledge-aware scenarios but poorly in others.
SKR bases retrieval on the model's knowledge of an answer, excelling in self-aware and time-aware scenarios but failing in others.
Additionally, since SKR uses BERT as the classifier,
whose internal knowledge has a significant gap with Llama,
it underperforms UAR with value heads based on the Llama's representation, in the self-aware scenario.

\subsection{Comparisons on Downstream Tasks}
For Self-RAG, we use inference scripts provided by the authors.
For FLARE, SKR, UAR, and always-retrieval methods, we use the same prompts to generate responses by incorporating the retrieved information.
We introduce the details of generation in Appendix \ref{apx:generation_details}.

The results are shown in Table~\ref{tab:main_result}. 
We see that UAR leads to the best overall performance across different downstream task scenarios, which indicates its effectiveness.
The percentage inside the parentheses represents the proportion of retrieval-invoked samples to the total samples.
We analyze each scenario as follows.

\paragraph{UAR does not invoke retrieval when factual knowledge is not needed.}
The DROP and GSM8K dataset do not require fact knowledge, and using retrieval enhancement will interfere with the model.
The results of always-retrieval are worse than never-retrieval. 
UAR only invokes a small amount of retrieval, while SKR and FLARE incorrectly invoke retrieval extensively.
And since UAR avoid unnecessary retrieval\footnote{UAR based on the 7B model incorrectly invokes retrieval 50\% of the time on the DROP dataset.
We speculate that this may be due to the limited representation capacity of the 7B model's hidden states.
In contrast, the 13B model only incorrectly invokes retrieval 3.7\% of the time.} and thus prevents affecting the original capabilities of the LLM,
it achieves the best results among all active retrieval methods on DROP and GSM8K, coming close to the results of never-retrieval.
Although Self-RAG does not incorrectly invoke retrieval, its final performance is not very good because it is fine-tuned based on the base model rather than leveraging the capabilities of the chat model.

\paragraph{UAR accurately invokes retrieval for time-sensitive questions.}
Since the questions in TAQA and FreshQA are time-sensitive and their answers keep changing, each question requires the retrieval of the latest information.
It is evident that the always-retrieval method based on Google Search performs significantly better than the never-retrieval method.
For TAQA, UAR almost perfectly invokes retrieval.
For FreshQA, UAR also invokes retrieval for most of the questions.
In contrast, other methods invoke retrieval less frequently and therefore do not use the latest information for responses, resulting in lower accuracy compared to UAR.

\paragraph{UAR accurately assesses the model's knowledge, avoiding poor retrieval impacts.}
For questions in TriviaQA and WQ whose answers do not change over time, always-retrieval is sub-optimal and the reason is two-fold: 1. For questions which model knows, retrieval increases unnecessary latency. 2. Potential incorrect external information will interfere correct internal knowledge.
Retrieving information only for knowledge that the model does not know can mitigate this issue.
Compared to SKR, UAR can more accurately determine whether the model knows a particular piece of knowledge.
Although SKR and UAR use a comparable number of retrieval calls, the accuracy of SKR's answers is lower than that of UAR, indicating that SKR's retrieval calls are less precise than UAR's.
We believe this is because SKR uses independent models, whereas our approach uses hidden states of the original model, resulting in better generalization.
Moreover, UAR outperforms always-retrieval with fewer retrieval calls, demonstrating the superiority of the Active Retrieval method.

\section{Analysis}

\subsection{Single Classifiers vs UAR}
\begin{table}[h]
\small
\centering
\begin{tabular}{lcc}
\toprule
\textbf{Scenario} & \textbf{Single Classifier} & \textbf{UAR}\\
\midrule
Intent-aware & 98.29 & 91.88 \\
Knowledge-aware & 99.66 & 90.38 \\
Time-aware & 99.41 & 86.69 \\
Self-aware & \underline{72.56} & \underline{72.32} \\
\bottomrule
\end{tabular}
\caption{Comparison between single classifiers and UAR based on Llama2-7B-chat.}
\label{tab:single_cls_acc}
\vskip -0.2in
\end{table}

Different scenarios have varying levels of discrimination difficulty.
As shown in Table \ref{tab:single_cls_acc}, the single classifier for the self-aware scenario has the lowest accuracy, which implies that determining whether the model is self-aware is a relatively challenging task.
We can also observe that the accuracy of each single classifier is higher than UAR in their respective scenarios.
The self-aware classifier may become the bottleneck restricting the performance of UAR, which also results in the accuracy of UAR on the AR-Bench being lower than the accuracy of using a single classifier alone.

\subsection{Using the Whole LLM as Classifier}
\begin{table}[h]
\centering
\small
\begin{tabular}{lcc}
\toprule
\textbf{Self-aware} & \textbf{Only Value Head} & \textbf{Whole LLM}\\
\midrule
Llama2-7B-chat & 72.56 & 75.65 \\
Llama2-13B-chat & 73.48 & 76.28 \\
\bottomrule
\end{tabular}
\caption{Comparison of the performance between training a value head as the classifier and training a entire large language model as the classifier.}
\label{tab:cls_vs_llm}
\vskip -0.2in
\end{table}

To improve the performance bottleneck of the self-aware classifier, we attempt to fine-tune the entire large language model as the classifier. 
From the results in Table \ref{tab:cls_vs_llm}, we can observe that on both 7B and 13B models, fine-tuning the entire model only achieves slight higher accuracy compared to just fine-tuning a lightweight value head.
Using a whole LLM as the classifier, UAR's inference latency and required parameters will significantly increase.
Therefore, we use lightweight value heads as classifiers, ensuring the efficiency of the entire framework with minimal performance loss.

\subsection{The Impact of Document Number}
\begin{figure}[h]
    \centering
    \includegraphics[width=\columnwidth]{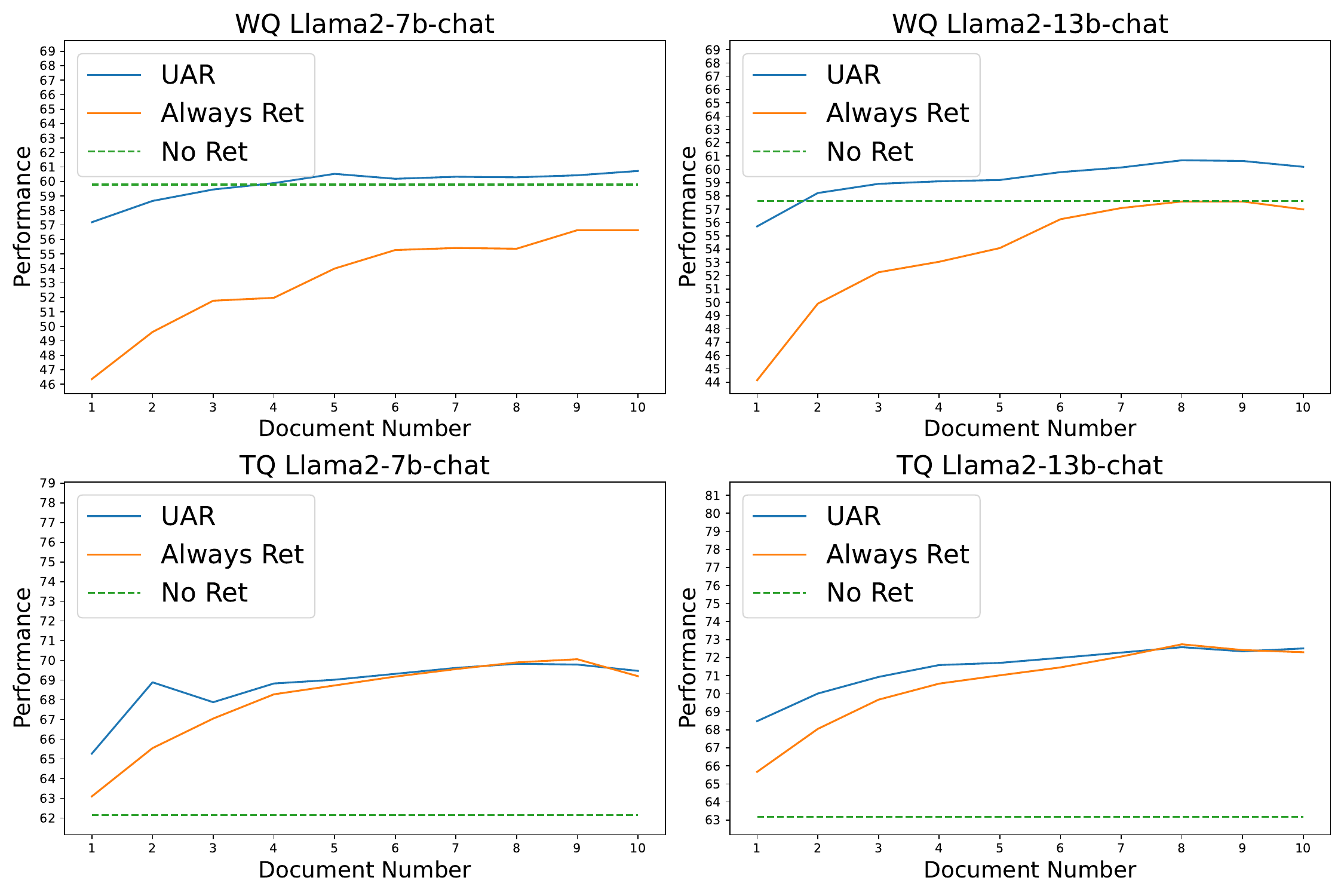}
    \caption{The impact of the number of reference documents on model performance.}
    \label{fig:doc_num}
    \vskip -0.2in
\end{figure}

We evaluate performance on the TriviaQA (TQ) and WebQuestions (WQ) datasets by varying the number of reference documents from 1 to 10.
The results, shown in Figure \ref{fig:doc_num}, indicate that on the WQ dataset, the always-retrieval method performs worse than the never-retrieval method, possibly because some documents disrupt the correct knowledge within the model.
UAR reduces retrieval frequency, enabling more precise retrieval calls and outperforming the never-retrieval method.
On the TQ dataset, always-retrieval outperforms never-retrieval, and performance improves with more documents, suggesting useful information might be in lower-ranked documents.
UAR performs best with fewer documents.
With more documents, it matches the performance of always-retrieval, although it requires significantly fewer retrieval calls.

\section{Conclusion}
In this paper, we introduce UAR, a unified active retrieval framework for retrieval-augmented generation.
Unlike existing methods that rely on a single criterion, UAR incorporates four orthogonal criteria into plug-and-play classification tasks, enabling comprehensive retrieval timing judgments with minimal inference cost and no loss of model capabilities.
We also introduce UAR-Criteria for processing various active retrieval scenarios uniformly.
We curate the Active Retrieval Benchmark (AR-Bench) to assess the retrieval timing accuracy of active retrieval methods across different scenarios.
Experimental results demonstrate that UAR significantly outperforms existing methods on AR-Bench and downstream tasks, highlighting its effectiveness and benefits to downstream applications.

\section*{Limitations}
We summarize limitations of our work as follows:
\begin{itemize}
    \item Our experiments primarily focus on the generation of short texts, such as in knowledge-based question answering, and involve only a single retrieval call.
    How to implement multiple active retrieval calls within longer text responses remains an area for future investigation.
    \item Our active retrieval criteria are primarily derived from our experience in practical applications, which may overlook some active retrieval scenarios.
    \item Our classifier is based on a single-layer MLP network. 
    Whether using a deeper network can further enhance performance remains to be explored.
\end{itemize}

\section*{Acknowledgement}
This work was supported by the National Natural Science Foundation of China (No. 62236004).  The computations in this research were performed using the CFFF platform of Fudan University.

\bibliography{custom}

\appendix
\section{Details of AR-Bench Construction}
\label{apx:ar_bench_details}

For the self-aware task, we employ the same method as described in Section \ref{sec:cls_training} to construct test samples on the TriviaQA validation set.
Questions the model does not know are marked as requiring retrieval.
The test set comprise 4000 questions the model knows and 4000 questions it does not.

For the time-aware task, we use 4000 time-sensitive questions from the TAQA test set as inputs requiring retrieval, and 4000 questions the model knows from the TriviaQA validation set as inputs not requiring retrieval.

For the knowledge-aware task, we use 4000 samples from the Self-RAG non-retrieval training data as inputs not requiring retrieval, and combine 2000 time-sensitive questions from the TAQA test set with 2000 questions the model does not know from the TriviaQA validation set as inputs requiring retrieval.

For the intent-aware task, we use 4000 questions the model knows from the TriviaQA validation set and 4000 instructions from the Self-RAG non-retrieval training data, half of which are concatenated with user retrieval intents as inputs requiring retrieval, and the other half as inputs not requiring retrieval.

It is important to note that the self-aware data for different models may vary, leading to different AR-Benches for different models.
In our experiments, we curate two separate AR-Benches for Llama2-7B-chat and Llama2-13B-chat respectively.

\section{Details of Baselines Re-implementation}
\label{apx:baseline_implementation}

\subsection{FLARE}
In implementing FLARE, we make two modifications.
First, we conduct experiments based on the Llama2-chat series of models, rather than using text-davinci-003.
Second, we eliminate the initial retrieval step in FLARE since our setting is active retrieval rather than passive retrieval.
We find that FLARE based on Llama2 struggle to achieve satisfactory results, which we suspect may be due to poor calibration of the Llama2-7B-chat and Llama2-13B-chat models.
The uncertainty estimation in FLARE heavily relies on model calibration, making it challenging to adapt to poorly calibrated models.
Therefore, on the AR-Bench, we conduct a direct search for the best retrieval thresholds for FLARE, ultimately setting them at 0.006 and 0.02 for the Llama2-7B-chat and Llama2-13B-chat models, respectively.

\subsection{SKR}

\begin{table}[h]
\small
\centering
\begin{tabular}{@{}lc@{}}
\toprule
\multicolumn{2}{c}{Training Hyper-parameters} \\ 
\midrule
Optimizer                    & AdamW \\
Warmup Steps                 & 0   \\
Learning Rate                & 2e-5 \\
Batch Size                   & 32   \\
Train Epochs                 & 5 \\
LR Scheduler                 & Linear   \\
Max-seq-length               & 512 \\
\bottomrule
\end{tabular}
\caption{Training hyper-parameters of SKR.}
\label{tab:skr_hyper_parameters}
\end{table}

In implementing SKR, we first use the 849 original pieces of data provided by the authors of SKR and collect self-knowledge data for the Llama2-7B-chat model according to the scripts in SKR's code repository. 
We obtain 15 questions that the model does not know and 143 questions that it knows, and find that these data are not sufficient to train an effective BERT classifier.
Therefore, we use the data from our training data of the self-aware classifier to train the BERT classifier for SKR.
Our training hyper-parameters are shown in Table \ref{tab:skr_hyper_parameters}.

\section{ChatGPT Evaluation}
\label{apx:chatgpt_eval}

We use gpt-3.5-turbo-instruct as the evaluator.
During the evaluation, we input the correct answer and the answer to be evaluated into gpt-3.5, and then let the model compare the correct answer with the answer to be evaluated to determine if the latter is correct.
Following \citet{iterative_ret}, we use the following prompt for evaluation.
{
\begin{lstlisting} [frame=none]
In the following task, you are given a Question, a model Prediction for the Question, and a Ground-truth Answer to the Question. You should decide whether the model Prediction implies the Ground-truth Answer.

Question:
{question}

Prediction:
{predicted answer}

Ground-truth Answer:
{ground-truth answer}
Does the Prediction imply the Ground-truth Answer? Output Yes or No:
\end{lstlisting}
}

\section{Details of Generation}
\label{apx:generation_details}

\subsection{Self-RAG}
We use the inference script provided by the Self-RAG authors for generation.
We determine the need for retrieval by whether the retrieval special token appears in the generated response.
For datasets using Contriever-MS MARCO as the retriever, we provide all 10 documents retrieved to Self-RAG for generation.

\subsection{Generation without Retrieval}
For the DROP dataset, we use the following prompt:
{
\begin{lstlisting} [frame=none]
Please answer the question based on the given passage. 
Passage: {passage in the dataset}
Question: {question}
Now give me the answer.
\end{lstlisting}
}

For the GSM8K dataset, we use the following prompt:
{
\begin{lstlisting} [frame=none]
Answer the math word question step by step. Your answer needs to end with 'The answer is'.
Question: {question}
Let's think step by step and give me the answer.
\end{lstlisting}
}

For other datasets, we directly input the question to the model:
{
\begin{lstlisting} [frame=none]
{question}
\end{lstlisting}
}

\subsection{Generation with Retrieval}
For the DROP dataset, we use the following prompt:
{
\begin{lstlisting} [frame=none]
Please answer the question based on the given passage. 
Passage: {passage in the dataset}
Question: {question}
    
Here are some additional reference passages:
{retrieved documents}

You can refer to the content of relevant reference passages to answer the questions.
Now give me the answer.
\end{lstlisting}
}

For the GSM8K dataset, we use the following prompt:
{
\begin{lstlisting} [frame=none]
Answer the math word question step by step. Your answer needs to end with 'The answer is'
Question: {question}
    
Here are some additional reference passages:
{retrieved documents}

You can refer to the content of relevant reference passages to answer the questions.
Let's think step by step and give me the answer.
\end{lstlisting}
}

For other datasets, we use the following prompt:
{
\begin{lstlisting} [frame=none]
{question}

Here are some additional reference passages:
{retrieved documents}

You can refer to the content of relevant reference passages to answer the questions.
Now give me the answer.
\end{lstlisting}
}

\section{Details of UAR Training}
\label{apx:uar_training}
When training the UAR classifiers, we set the batch size to 32 and train for a total of 10 epochs, saving after each epoch and selecting the checkpoint that perform best on the validation set.
We conduct a grid search on the validation set and ultimately determine the learning rate to be 5e-5.
Our classifier is a fully connected layer with an input dimension equal to the hidden state dimension and an output dimension of 2.

\section{Downstream Task Datasets}
\label{apx:downstream_dataset}

For knowledge-aware scenario, we use the validation set of DROP \citep{DROP} and the test set of GSM8K \citep{GSM8K} as the test sets.
DROP is a reading comprehension benchmark, which needs the model to answer questions based on given paragraphs.
GSM8K is a dataset containing diverse grade school math word problems, primarily used to assess the reasoning ability of models.
These two datasets evaluate the model's abstract abilities, e.g., reading comprehension and math reasoning, and thus do not require extra fact knowledge.
Therefore, they can measure the ability of active retrieval methods to avoid unnecessary retrieval for scenarios that requires little fact knowledge.

For time-aware scenario, we use the test set of TAQA \citep{TAQA} and questions whose answers will change over time from FreshQA \citep{FreshQA} (We remove questions with false premises).
Since these questions are time-sensitive, the active retrieval system need to retrieve real-time information for every question.

For self-aware scenario, we use the validation set of TriviaQA \citep{TriviaQA} and the test set of WebQuestions (WQ) \citep{WQ}.
These test samples are non-time-sensitive questions.
The active retrieval system only needs to retrieve questions which the model does not know, and try to achieve high answer accuracy with an appropriate number of retrieval calls.

\end{document}